\documentclass{article}

\PassOptionsToPackage{square, numbers, sort&compress}{natbib}
\usepackage[preprint]{tackling_climate_workshop_style}

\usepackage[utf8]{inputenc} % allow utf-8 input
\usepackage[T1]{fontenc}    % use 8-bit T1 fonts
\usepackage{hyperref}       % hyperlinks
\usepackage{url}            % simple URL typesetting
\usepackage{booktabs}       % professional-quality tables
\usepackage{amsfonts}       % blackboard math symbols
\usepackage{nicefrac}       % compact symbols for 1/2, etc.
\usepackage{microtype}      % microtypography
\usepackage{amssymb}
\usepackage{amsmath}
\usepackage{mathtools}
\usepackage{siunitx}
\usepackage{subcaption}
\usepackage{float}
\usepackage{siunitx}
\usepackage[export]{adjustbox}

\usepackage{placeins}
\usepackage{appendix}

\title{Sim2Real for Environmental Neural Processes}

\author{%
   Jonas Scholz \\
   University of Cambridge \\
   Cambridge, UK \\
   \texttt{js2731@cam.ac.uk} \\
   \And
   Tom R. Andersson \\
   British Antarctic Survey \\
   Cambridge, UK \\
   \texttt{tomand@bas.ac.uk} \\
   \AND
   Anna Vaughan \\
   University of Cambridge \\
   Cambridge, UK \\
   \texttt{av555@cam.ac.uk} \\
   \And
   James Requeima \\
   Vector Institute, \\
   Toronto, ON, Canada \\
   \small{\texttt{james.requeima@vectorinstitute.ai}} \\
   \And
   Richard E. Turner \\
   University of Cambridge \\
   \& Microsoft Research \\
   \texttt{ret26@cam.ac.uk} \\
}

\begin{document}

\maketitle

\begin{abstract}
Machine learning (ML)-based weather models have recently undergone rapid improvements.
These models are typically trained on gridded reanalysis data from numerical data assimilation systems.
However, reanalysis data comes with limitations, such as assumptions about physical laws and low spatiotemporal resolution.
The gap between reanalysis and reality has sparked growing interest in training ML models directly on observations such as weather stations.
Modelling scattered and sparse environmental observations requires scalable and flexible ML architectures, one of which is the convolutional conditional neural process (ConvCNP).
ConvCNPs can learn to condition on both gridded and off-the-grid context data to make uncertainty-aware predictions at target locations.
However, the sparsity of real observations presents a challenge for data-hungry deep learning models like the ConvCNP.
One potential solution is `Sim2Real': pre-training on reanalysis and fine-tuning on observational data.
We analyse Sim2Real with a ConvCNP trained to interpolate surface air temperature over Germany, using varying numbers of weather stations for fine-tuning.
On held-out weather stations, Sim2Real training substantially outperforms the same model architecture trained only with reanalysis data or only with station data, showing that reanalysis data can serve as a stepping stone for learning from real observations.
Sim2Real could thus enable more accurate models for weather prediction and climate monitoring.
\end{abstract}

\vspace{5mm}
\section{Introduction}
Every day, millions of observations of the Earth system are collected by environmental sensors, such as in-situ weather stations, satellites, aircraft, oceanographic buoys, and meteorological balloons  \citep{hersbach2020era5}.
However, the spatial distance between neighbouring observations can be very large for certain variables, particularly in remote regions like Antarctica or the Himalayas.
This presents a challenge for flexible deep learning models that require an abundance of data to learn realistic physical behaviour.
As a result, deep learning systems developed for environmental prediction tasks often use gridded \textit{reanalysis} datasets for training, rather than observations.
Reanalysis data is produced by assimilating observations into a dynamical model to predict geophysical quantities (such as temperature) on a regular spatiotemporal grid \citep{hersbach2020era5}.
Data-driven models trained with reanalysis data have recently made striking progress in weather forecasting \cite{lam2022graphcast, nguyen2023climax, chen2023fengwu, pathak2022fourcastnet, bi2023accurate}.

A limitation of training with reanalysis target data is the mismatch between the dynamical model simulator and reality, owing to several factors, including:
\begin{itemize}
\addtolength\itemsep{-1.5mm}
    \item error between physical laws in the simulator and true physics,
    \item spatiotemporal coarseness (typically $\sim \SI{25}{km}$ spatially and $\sim \SI{1}{hr}$ temporally),
    \item not capturing real-world observation aleatoric uncertainty (e.g. sensor noise).
\end{itemize}
We call the combination of the above mismatches the `Sim2Real gap'.
Despite rapid advances in reanalysis-trained large machine learning (ML) weather forecasting models, such as GraphCast \citep{lam2022graphcast} and Pangu-Weather \citep{bi2023accurate}, the Sim2Real gap will degrade the utility of these models to some degree.
A recent approach to overcoming the Sim2Real gap in weather forecasting is MetNet-3 \citep{andrychowicz2023metnet3}, which forecasts both simulator and observation data in a multi-task setting.
An important open question is whether ML weather models can be trained purely on environmental observations, perhaps bypassing the data assimilation step of numerical simulators and overcoming the Sim2Real gap.

\begin{figure}
    \centering
    \includegraphics[width=\textwidth]{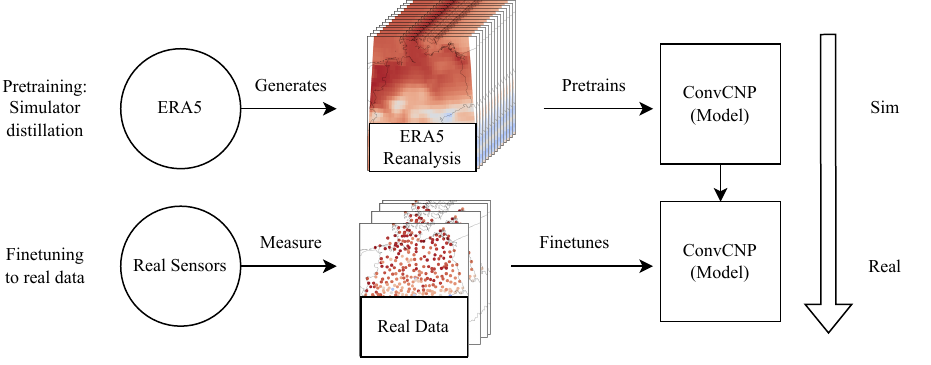}
    \caption{In the Sim2Real process, a ConvCNP is pre-trained on abundant simulator data from the ERA5 reanalysis dataset, and then fine-tuned on limited real weather station data.}
    \label{fig: intro sim2real}
\end{figure}

In this paper, we investigate Sim2Real transfer for convolutional conditional neural processes (ConvCNPs) 
\citep{gordon2019convolutional}.
ConvCNPs are flexible deep learning models that produce probabilistic predictions over $\mathbf{y}_T$ at target locations $X_T$ conditioned on context observations $(X_C, \mathbf{y}_C)$ (Appendix \ref{apx.arch_and_experiment}). Convolutional neural process variants have shown promising results in weather modelling tasks such as downscaling \citep{vaughan2022convolutional} and sensor placement \citep{andersson2022active}.
However, the model in \citep{andersson2022active} was trained with reanalysis data as input and output.
The Sim2Real gap motivates training a ConvCNP to interpolate real station data, which could provide more realistic estimates of observation informativeness and suggest better sensor placements than a reanalysis-only ConvCNP.
In this work, to quantify the benefits of Sim2Real, a ConvCNP is first pre-trained on vast amounts of reanalysis data, and then a smaller but higher-quality real weather station dataset is used to fine-tune and evaluate the model (Fig. \ref{fig: intro sim2real}).

\section{Spatial Temperature Interpolation}

In this experiment, a ConvCNP is trained to spatially interpolate 2-metre air temperature context observations at a given time instant $(X_{C}, \mathbf{y}_{C})$ to predict target values $\mathbf{y}_{T}$ at the same time instant at arbitrary target locations $X_T$.
A Gaussian likelihood is used with a negative log-likelihood (NLL) loss.
ERA5 reanalysis \citep{hersbach2020era5} is used as the simulator data and weather stations from the German weather service (DWD) \citep{dwd} are used as the real data.
To aid predictions, the ConvCNP receives a second context set with high-resolution elevation data from the NASA Shuttle Radar Topography Mission \citep{farr2007shuttle}, as well as the time of day, day of year, and normalised latitude/longitude.
These auxiliary features enable the model to learn spatiotemporal non-stationarities in the data.
Further details on the ConvCNP architecture and training procedure are provided in Appendix \ref{apx.arch_and_experiment}.

During pre-training, tasks are generated by randomly sampling ERA5 grid cells for the context and target.
During fine-tuning, off-the-grid DWD weather station observations are randomly split into context and target.
53 DWD stations are held-out during training/validation and are used to evaluate the generalisation abilities of the trained models.
The held-out stations are taken from the VALUE protocol, which carefully selects stations that cover a wide range of geographic environments \citep{maraun2015value} (Appendix \ref{apx.data_splitting}).
Different data-availability regimes are analysed, both spatially in terms of the number of stations available for training/validation $N_{stations} \in \{ 20, 100, 500 \}$ and temporally in terms of the number of time slices used for training $N_{times} \in \{16, 80, 400, 2000, 10000\}$. 
Two baselines are used: `Sim Only' (training the ConvCNP only to interpolate ERA5 and applying it directly to DWD data) and `Real Only' (training the ConvCNP only to interpolate the DWD data).
Different Sim2Real adaptation methods were investigated, mostly focusing on global fine-tuning of all parameters, and FiLM adaptation \citep{perez2018film} (Appendix \ref{apx.finetuning}).

\section{Results \label{sec: temperature results}}
\begin{figure}
    \centering
    \includegraphics[width=\textwidth]{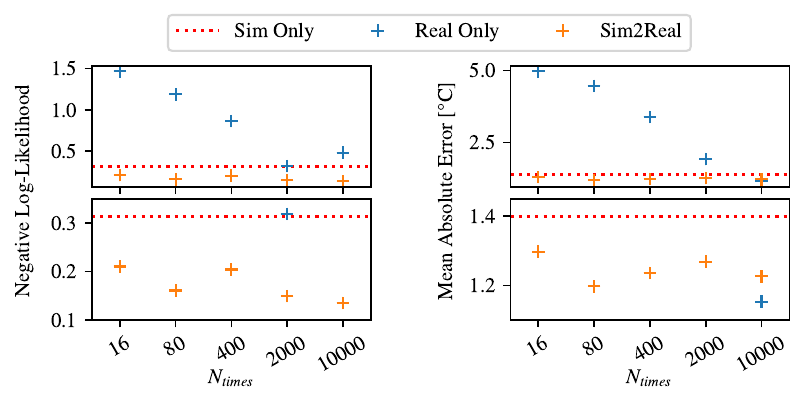}
    \caption{Sim2Real outperforms Real Only and Sim Only baselines significantly when $N_{stations} = 500$. The bottom row is a zoomed-in version of the top row.}
    \label{fig: sim2real results}
\end{figure}

Global fine-tuning substantially outperforms FiLM adaptation in the ERA5 $\rightarrow$ DWD Sim2Real experiment.
To adapt to higher-frequency correlations, we hypothesise the adaptation strategy must update the ConvCNP's convolutional filters (the smallest DWD station separation is $\sim 5\times$ smaller than the ERA5 grid spacing).
FiLM adaptation does not update the CNN weights, so it struggles to adapt to shorter length scale features in the real data.
This hypothesis is validated in a separate Sim2Real toy experiment using 1D Gaussian processes (Appendix \ref{apx.synthetic_experiment}, Fig. \ref{fig: shrinking l results}).

Using global fine-tuning, Sim2Real produces substantially better NLLs (left) and mean absolute errors (MAEs) (right) than the Sim Only and Real Only baseline when there is a sufficiently large number and density of training stations ($N_{stations} = 500$).
Sim2Real outperforms the Sim Only baseline even with a very small temporal window of station data ($N_{times} = 16$).
However, the benefit of Sim2Real decreases as real data becomes more temporally abundant; the Real Only baseline's performance is comparable to that of Sim2Real at $N_{times} = 10000$ (Fig. \ref{fig: sim2real results}).
Furthermore, in the sparser station settings ($N_{stations} \in \{20, 100\}$), the fine-tuning stage of Sim2Real does not outperform the Sim Only pre-trained baseline as clearly (Appendix \ref{apx.full_experiment}).

The results in Figure \ref{fig: sim2real results} are noisy because only a single model was trained in each condition, and due to mismatches between testing and validation stations leading to poor early-stopping (Appendix \ref{apx.data_splitting}).
Future work should average the results over more training runs with different random initialisations.

\paragraph{Sim2Real Learns High-Frequency Features \label{sec: temperature high frequency features}}
\begin{figure}
    \centering
    \includegraphics[width=0.9\textwidth]{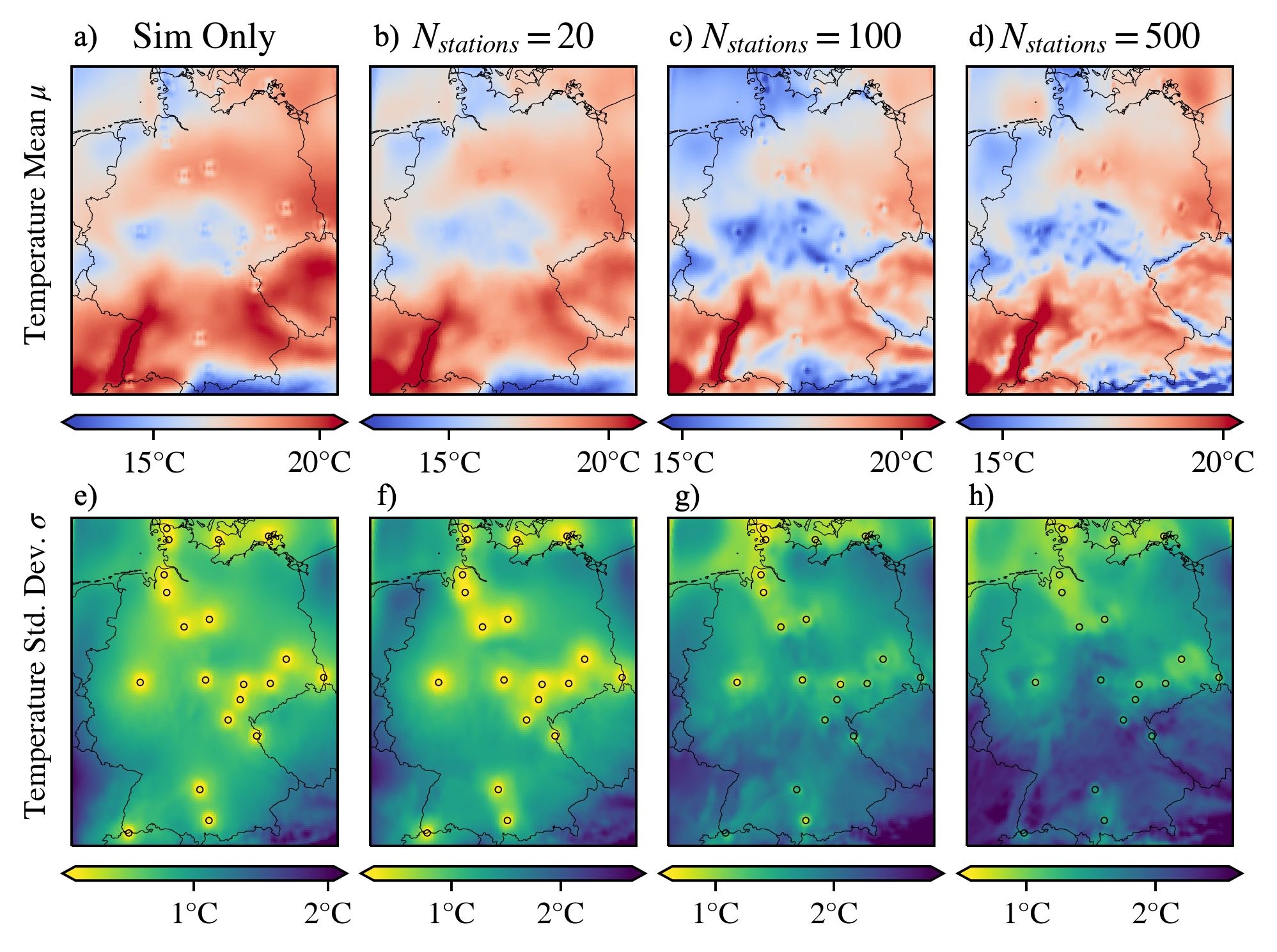}
    \caption{Increasing the density of training stations for fine-tuning increases the resolution of the ConvCNP's  predictions. \textit{a--d}, ConvCNP mean, $\mu$. \textit{e--h}, ConvCNP standard deviation, $\sigma$. Context weather stations are shown as black circles (not shown in \textit{a--d} to reveal the spatial artefacts in $\mu$).}
    \label{fig: superresolution}
\end{figure}

When the ConvCNP is trained on ERA5, the shortest learnable correlation lengthscale is the grid spacing $\ell_{min}^{sim}$. Fine-tuning the model on real data shortens $\ell_{min}$ to (roughly) the shortest inter-station separation, $\ell_{min}^{real}$, which is a factor of 5 smaller than $\ell_{min}^{sim}$ for the densest DWD station setting of $N_{stations} = 500$. This leads to higher-frequency features in the model's predictions, modelling shorter-lengthscale weather phenomena (Fig. \ref{fig: superresolution} a-d).
The Sim Only ConvCNP produces overconfident marginal predictive uncertainties $\sigma$ around context observations (Fig. \ref{fig: superresolution} e).
As $N_{stations}$ is increased, the model learns more appropriate conditioning behaviour for real observational data: in regions with multiple consistent measurements, uncertainty is reduced in a wide region (e.g. Northern Germany in Fig. \ref{fig: superresolution} h), if nearby observations disagree, the Sim2Real ConvCNP inflates its uncertainty (e.g. central Germany).
Furthermore, the average predictive uncertainties are greater.
Even though Sim2Real yields these non-trivial predictive improvements, it is unable to remove the unphysical visual artefacts around context observations in Fig. \ref{fig: superresolution} a-d, which we hypothesise is caused by the lack of training signal below $\ell_{min}$ (Appendix \ref{apx.artefacts}).

\section{Discussion}
This paper explores Sim2Real with a ConvCNP in a \SI{2}{m} temperature interpolation task over Germany, transferring from ERA5 reanalysis to real weather station observations.
Our preliminary experiments paint a picture where Sim2Real is highly effective in a `medium-data' regime.
Too little real data and fine-tuning the pre-trained model has minimal effect (with the spatial data abundance being more important than the temporal abundance).
Plenty of real data and the pre-training phase loses its value; starting from random initialisation is just as effective.
Where Sim2Real is effective, it could bridge gaps in the finite observational data, improving downstream model usage such as active learning for sensor placement \citep{andersson2022active}.
In future work, transfer learning from data-rich areas like Germany to data-sparse areas like the Himalayas or Antarctica could further alleviate data gaps and address socioeconomic disparities.

Our work identifies a limitation of ConvCNPs when training with spatially sparse data: the model can only make robust predictions in the range of length scales featured within the data.
In particular, prediction artefacts appear on length scales shorter than the shortest context-target separation during training.
This is likely the case for other ML models based on CNNs.
Future work should explore architectural or training approaches to alleviate this.

We expect that Sim2Real can contribute to the rapidly developing future of data-driven weather and climate modelling.

\newpage

\begin{ack}
Code is made available at \url{https://github.com/jonas-scholz123/sim2real-downscaling}. This paper builds heavily on the \texttt{DeepSensor} \citep{deepsensor} and \texttt{neuralprocesses} \citep{neuralprocesses} packages.

We thank the two anonymous reviewers from the CCAI NeurIPS 2023 workshop for their helpful feedback.

R.E.T. is supported by Google, Amazon, ARM, Improbable and EPSRC grant EP/T005386/1.

T.R.A is supported by Wave 1 of The UKRI Strategic Priorities Fund under the EPSRC Grant EP/W006022/1, particularly the AI for Science theme within that grant and The Alan Turing Institute.

Anna Vaughan acknowledges the UKRI Centre for Doctoral Training in the Application of Artificial Intelligence to the study of Environmental Risks (AI4ER), led by the University of Cambridge and British Antarctic Survey, and studentship funding from Google DeepMind.
\end{ack}

\bibliography{bibliography}

%%%%%%%%%%%%%%%%%%%%%%%%%%%%%%%%%%%%% APPENDIX

\newpage
\setcounter{page}{1}
\begin{appendix}

\newcommand{\appendixhead}%
{\noindent \textbf{\Large Appendix}}

\begin{center}
    \appendixhead
    \vspace{0.1in}
\end{center}

\vspace{0.25in}

% 

% Make figure/table counters restart within sections for the appendix, and prepend the appendix letter when referencing
\counterwithin{figure}{section}
\counterwithin{table}{section}
\counterwithin{equation}{section}
\renewcommand\thefigure{\thesection\arabic{figure}}
\renewcommand\thetable{\thesection\arabic{table}}

% \hl{TODO finish updating appendix.}
\FloatBarrier
\section{Finetuning Approaches}\label{apx.finetuning}

\paragraph{Global Finetuning}
In \textit{global} finetuning, all parameters of the pretrained model are trained on the finetuning dataset. Tuning all parameters gives the greatest degree of flexibility, but the high capacity also makes the model prone to overfitting, particularly if the finetuning dataset is small.

\paragraph{FiLM Adapters}
To reduce the model's susceptibility to overfit to small finetuning datasets, some parameters are commonly frozen and the remaining parameters are trained on the finetuning data. FiLM adaptation \citep{perez2018film} is one data-efficient approach that has shown strong results within computer vision \cite{perez2018film, gupta2022towards}.

FiLM adapters are affine transformations applied to individual feature maps $\mathbf{h}^{(l)}_i$ throughout the model:
\begin{equation}
    \tilde{\mathbf{h}}^{(l)}_i = \gamma^{(l)}_i \times \mathbf{h}^{(l)}_i + \beta^{(l)}_i,
\end{equation}
where $\gamma^{(l)}_i, \beta^{(l)}_i$ are the learnable FiLM parameters scaling and shifting the $i$th feature map of the $l$th layer, $i \in 1\dots N_{feat}^{(l)}$. Because a FiLM layer contains only $2N_{feat}$ trainable parameters, it is relatively robust to overfitting. We apply FiLM adapters after every convolutional layer as shown in Fig. \ref{fig: model arch} (orange). During pre-training, we fix $\beta_i^{(l)} = 0, \gamma_i^{(l)} = 1$. During finetuning, we then freeze all other model parameters and train only the FiLM parameters.

\FloatBarrier
\section{Synthetic Experiment: 1D Gaussian Progress Regression}\label{apx.synthetic_experiment}
In this synthetic experiment, we perform ``Sim2Real" by generating both the ``simulated" and the ``real" data, both drawn from a noisy 1D Gaussian Process (GP) \citep{williams2006gaussian}. The GP uses a squared-exponential kernel with lengthscales $\ell^{sim}, \ell^{real}$ and noise $\sigma_0^{sim}, \sigma_0^{real}$ for the ``sim" and ``real" data respectively. This experiment served as a stepping stone for rapid iteration and as a preliminary evaluation of adaptation methods, which partially translate to the main temperature downscaling experiments.

\paragraph{Baselines}
As baselines, we consider the ConvCNP trained in the ``infinite" data regime, where we keep training it until convergence, and the 0-shot ``Sim Only" baseline, where we apply the simulator-pretrained ConvCNP directly to the ``real" data.

\paragraph{Shrinking Lengthscales}
In the first experiment, we keep noise fixed at $\sigma_0^{real} = \sigma_0^{sim} = 0.05$ and consider the transfer from $\ell^{sim} = 0.25$ to \textit{shorter} lengthscale GPs, with $\ell^{real}$ of either 0.2, 0.1 or 0.05. This is analogous to the target domain of weather modelling, where some real measurement stations are closely separated ($ \ell^{real} \sim \SI{4}{km}$) and can therefore capture shorter lengthscale weather phenomena than the more coarsely gridded ERA5 simulator data with $\ell^{sim} \sim \SI{20}{km}$ grid spacing.\footnote{The analogy is not perfect, as long-lengthscale weather phenomena are still present in both real and simulator data, which is not the case for these GPs, but a part of the problem is captured nonetheless.}

\begin{figure}
    \centering
    \includegraphics[width=\textwidth]{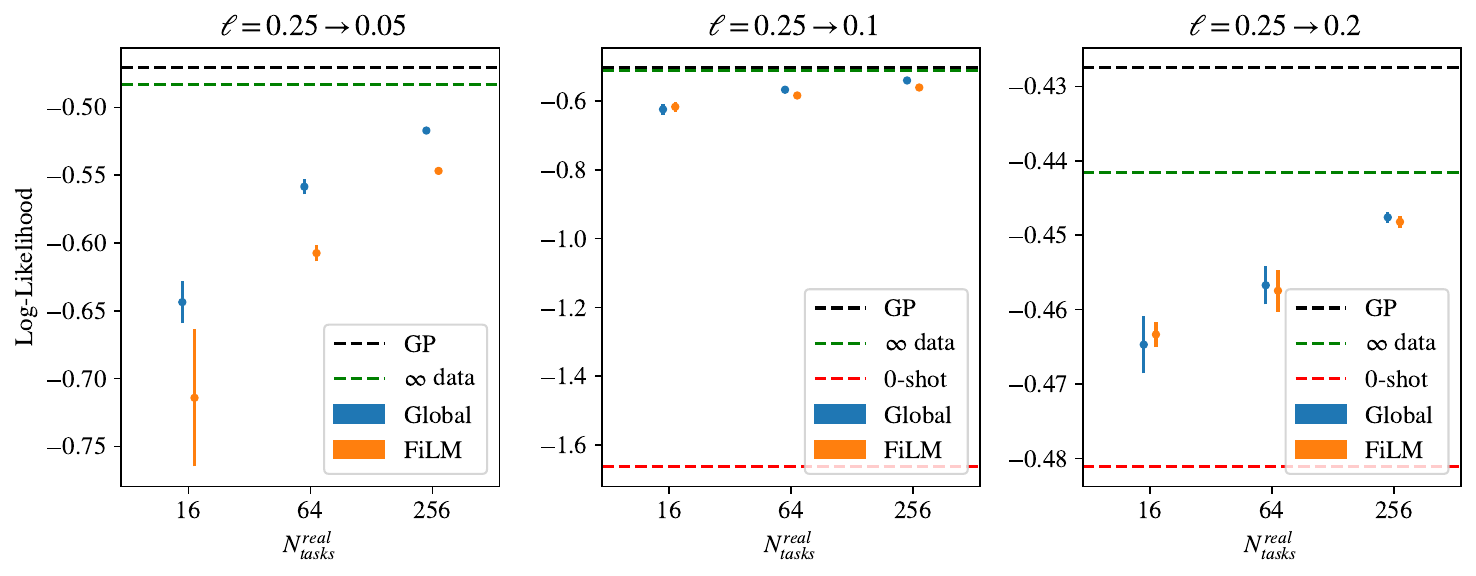}
    \caption{Test-set log-likelihoods achieved via fine-tuning on limited numbers of ``real" tasks (x-axis). Global fine-tuning outperforms FiLM in the shrinking lengthscale experiments, particularly if the difference in lengthscales is large. FiLM performs slightly worse the more fine-tuning data are available. Error bars represent 95\% confidence intervals and are computed by starting from the same pre-trained model and using different fine-tuning datasets. The 0-shot baseline in the left-most plot is $\approx -4.1$ and is hidden to not distort the y-scale.}
    \label{fig: shrinking l results}
\end{figure}

As shown in Fig. \ref{fig: shrinking l results}, both FiLM and global fine-tuning require only a very small number of tasks for effective adaptation to shorter lengthscales, when compared to the (poor) 0-shot baseline. In the more extreme $\ell = 0.25 \to 0.05$ transfer, global fine-tuning significantly outperforms FiLM. We hypothesize that the convolutional filters learned on the pre-training task extract features that are tuned to the particular $\ell^{sim} = 0.25$ -- for less extreme changes in $\ell$, FiLM adaptation is able to scale features to achieve similar performance to global fine-tuning, for much smaller lengthscales, the features extracted by the convolutional filters become less useful and fine-tuning the filters themselves becomes important.

Overall, this effect is smaller for smaller values of $N_{tasks}^{real}$. In these sparse-data regimes, the much lower capacity of FiLM adaptation makes the model less likely to overfit the ``real" data. For $\ell^{real} \in (0.1, 0.2)$, and $N_{tasks}^{real} = 16$, this leads to FiLM slightly outperforming global finetuning.

\paragraph{Growing Lengthscales}
\begin{figure}
    \centering
    \includegraphics[width=\textwidth]{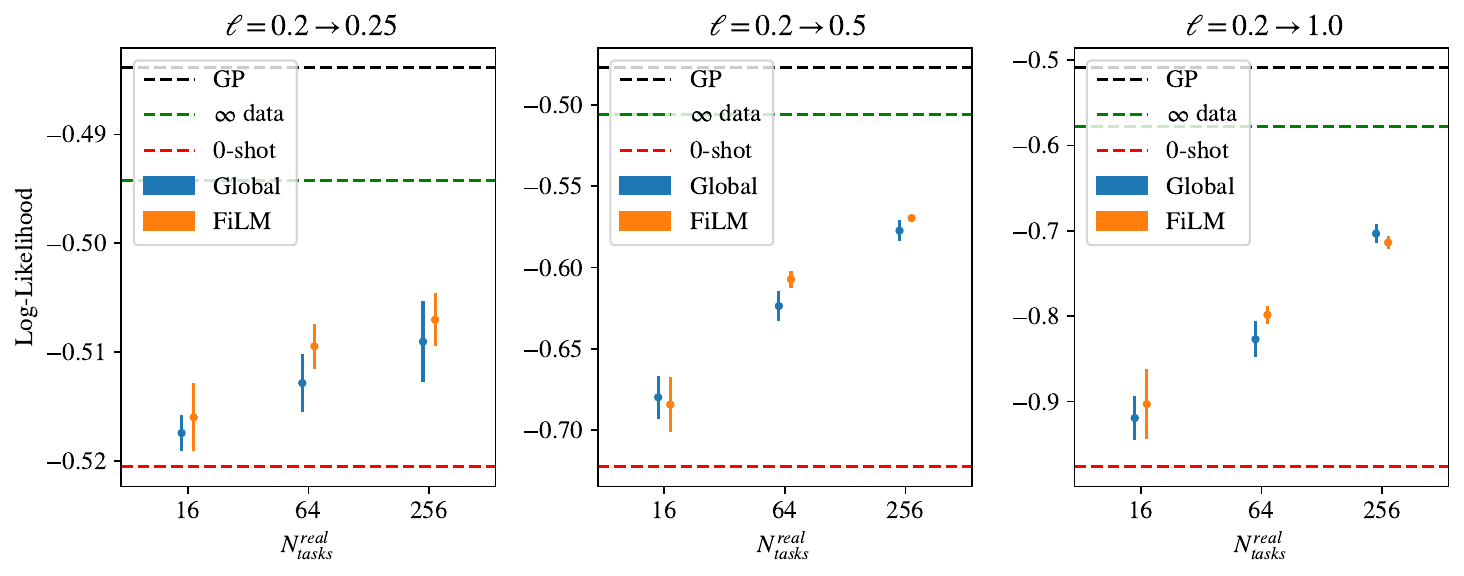}
    \caption{For growing lengthscales, FiLM outperforms global fine-tuning. Especially for sparse data settings and small Sim2Real gaps.}
    \label{fig: gp growing lengthscales}
\end{figure}

If our hypothesis that FiLM layers cannot rectify low-resolution convolutional filters holds, we should see an improved FiLM performance in the inverse problem of growing lengthscales. This setting is less analogous to the real weather modelling experiment but is useful to include for generality and a more domain-agnostic approach to fine-tuning.

We therefore now consider $\ell^{sim} = 0.2$ and $\ell^{real} \in [0.25, 0.5, 1.0]$. As hypothesised, FiLM performs slightly better in this setting (Fig. \ref{fig: gp growing lengthscales}), particularly in the sparse-data regime.
%However, the benefit of one method over another should not be overstated, especially considering the substantial standard deviation in performance associated with different fine-tuning datasets.

\paragraph{Noise Change}
\begin{figure}
    \centering
    \includegraphics[width=\textwidth]{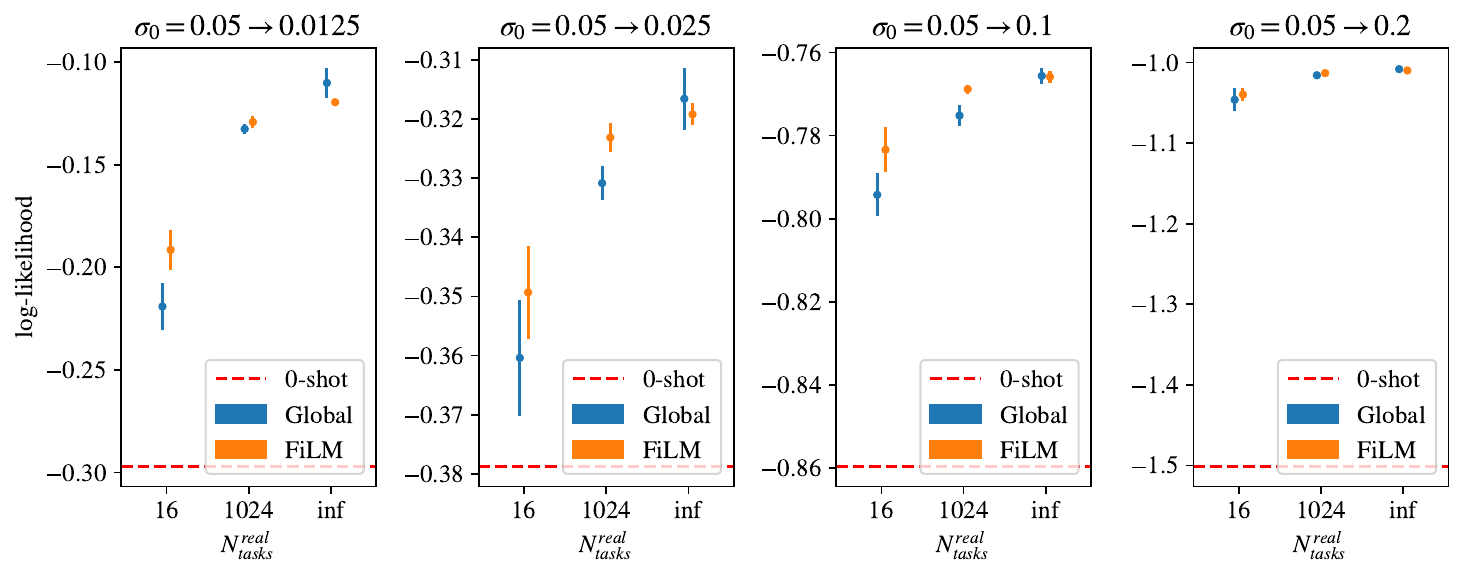}
    \caption{When adapting to different noise levels, FiLM adapters beat global fine-tuning decisively. Even in large data regimes, FiLM is only very slightly weaker.}
    \label{fig: gp noise change results}
\end{figure}

Finally, we keep the lengthscales fixed at $\ell^{sim} = \ell^{real} = 0.25$, and instead change the level of \textit{noise} from $\sigma_0^{sim} = 0.05$ both up to $\sigma_0^{real} \in [0.1, 0.2]$ and down to $\sigma_0^{real} \in [0.0125, 0.025]$.

This is analogous to our weather-based Sim2Real experiments, where we would expect our simulated data, ERA5 \citep{hersbach2020era5}, to be associated with lower noise\footnote{By noise we do not mean (negligible) inaccuracies in the temperature measurement, but instead the aleatoric uncertainty due local weather phenomena (e.g. a cloud flying overhead at the time of measurement or a cold breeze passing by) that might lead to temperature changes on the order of seconds and metres, which are infeasible to model.} than real measurements because
\begin{itemize}
    \item The simulation runs on a discrete spatiotemporal grid and therefore cannot model weather phenomena beyond its resolution.
    \item The data assimilation process can ingest multiple data points of observational data within one grid cell, smoothing out local measurement noise.  
\end{itemize}

For generality, we both simulate an increase and a reduction in noise.

In Fig. \ref{fig: gp noise change results}, we show that in this regime FiLM outperforms global fine-tuning across different values of $\sigma_0^{real}$, even in the larger $N_{tasks}^{real} = 1024$ experiments. In the limit $N_{tasks}^{real} \to \infty$, global fine-tuning still outperforms (as it should), but not by a large margin.

These results align with our previous findings, that FiLM is a more sample-efficient fine-tuning method \textit{unless} the frozen convolutional filters extract features of an insufficient resolution.

\section{Model Architecture and Experimental Details}\label{apx.arch_and_experiment}
\begin{figure}
    \centering
    \includegraphics[width=\textwidth]{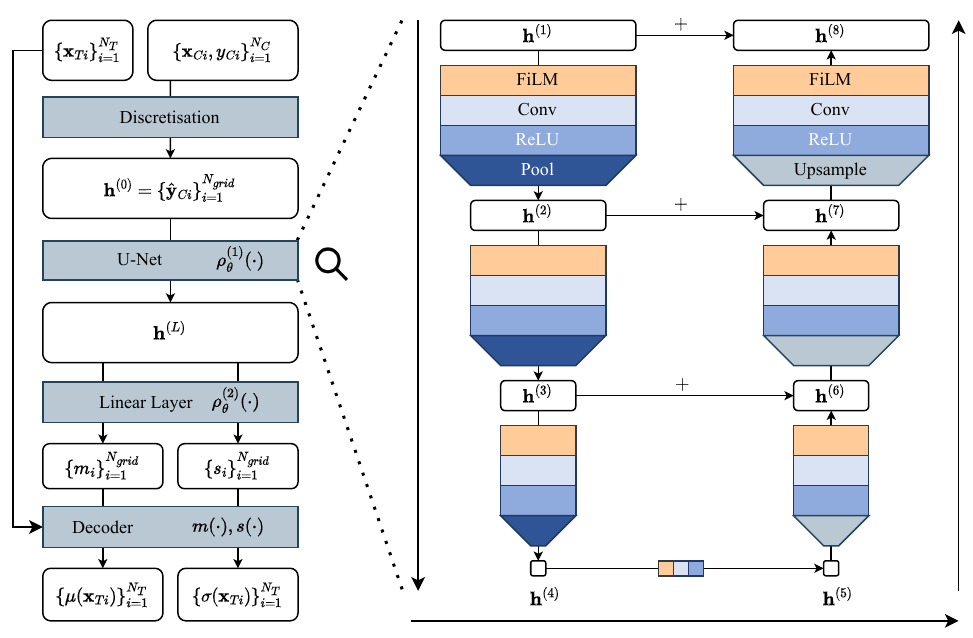}
    \caption{The ConvCNP architecture, using a U-Net \citep{ronneberger2015unet} (and linear layer) as $\rho_\theta$. Coloured boxes with sharp corners are the processing steps and rounded white boxes are the different states along the processing pipeline.}
    \label{fig: model arch}
\end{figure}

\paragraph{Convolutional Conditional Neural Processes}
ConvCNPs \citep{gordon2019convolutional} are spatiotemporal models with parameters $\theta$ that define the conditional distribution over target variables $\mathbf{y}_T$ at given target locations $X_T$ as a Gaussian distribution
\begin{equation}
 q_\theta(\mathbf{y}_T | X_T, C) = \mathcal{N}(\mathbf{y}_T; \boldsymbol{\mu}_\theta(C), \Sigma_\theta(C)).
\end{equation}

They model $\boldsymbol{\mu}$ and $\Sigma$ by encoding context data $C = \{\mathbf{x}_{Ci}, y_{Ci}\}_{i = 1}^{N_C}$, onto an internal gridded representation, processing that encoding into a (gridded) predictive mean and standard deviation using a Convolutional Neural Network (CNN) \citep{cnns} $\rho_\theta$, and finally decoding the gridded mean and standard deviation at any on or off-the-grid target locations. It is this CNN $\rho_\theta$ that we fine-tune in our experiments. The particular architecture we use is a 
U-Net \citep{ronneberger2015unet}, shown in Fig. \ref{fig: model arch}.

ConvCNPs are trained by adjusting the parameters $\theta$ to minimise the Negative Log-Likelihood (NLL) of the predictive distribution over the training set, via backpropagation, to minimise the KL-Divergence between the approximate posterior predictive $q_\theta$ and the \textit{true} posterior predictive.

\paragraph{Normalisation}
We normalise input data so that each of the two spatial coordinates is normalised to the range [0, 1]. Additionally, we normalise temperatures by subtracting their sample mean and scaling by their sample standard deviation. We save normalisation parameters during pre-training and use them during fine-tuning for consistency. This is performed using the \textit{deepsensor} package \citep{deepsensor}.

\paragraph{Model Hyperparameters}
We use the model architecture shown in Fig. \ref{fig: model arch} with 6 layers (down and up) in the U-Net, of 96 channels each. We choose a resolution of 200 Points Per Unit\footnote{Note that this means $200 \times 200$ grid-points per $1 \times 1$ unit square} (PPU) for the internal gridded representation and a corresponding encoder and decoder lengthscale $\ell_E = \ell_D = 1/200$, which allows us to comfortably resolve the smallest station separation ($\sim \SI{4}{km}$) in normalised space. In total, our model has 3.8 million parameters, out of which 3284 (0.08\%) are FiLM parameters.

\paragraph{Optimisation}
We use the Adam optimiser \citep{kingma2014adam}, with a learning rate of $\num{1e-4}$ during pre-training and $\num{3e-5}$ during fine-tuning, both of which were found via grid search. We use a batch size of 16 throughout. Because each point in time yields a very large combination of context and target-set combinations (which are related but distinct) an ``epoch" in the traditional sense is far too large to be useful. We instead define an epoch as 200 batches (i.e. $200 \times 16 = 3200$ tasks) during pre-training. We anneal the learning rate by a factor of 3 if the validation loss stalls for more than 8 epochs, which helps for convergence at the end of training. We stop after 20 epochs without improvement. During fine-tuning, we define epochs to be smaller, each consisting of 25 batches, to monitor validation losses more frequently. This definition of epoch also allows for consistency across different $N_{times}$ and $N_{stations}$. During fine-tuning, we stop after 30 such ``epochs" without improvement.

\FloatBarrier
\section{Data Splitting and Sampling}\label{apx.data_splitting}
\begin{figure}
    \centering
    \includegraphics[width=1\textwidth, center]{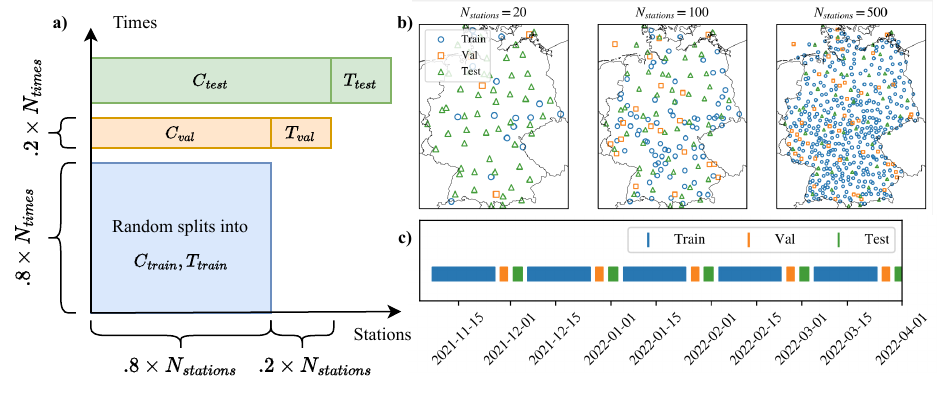}
    \caption{a) The $N_{times}$ times and $N_{stations}$ stations each get split into 80\% training and 20\% validation data. For validation, we use training stations as context. For final testing, we use all available stations (train and val) to achieve the best results. The set of \textit{test} times and stations is set aside at the beginning and is used for all evaluations. b) Train/Val/Test stations for different $N_{stations}$. c) Train/Val/Test dates are sampled throughout the available period with 2 days discarded between to avoid leakage.}
    \label{fig: data splitting}
\end{figure}

Spatiotemporal modelling makes splitting data significantly more complex than it is in most traditional ML domains. Generating training, validation and testing sets is not as simple as splitting all available data randomly. The most important problem is that the model can overfit both spatially and temporally -- both of which can leak into the test/validation sets unnoticed unless care is taken.

The data-splitting process is further complicated by the fact that we're exploring different data splits to investigate different data-availability regimes.

\paragraph{Test Data}
We split the available real data along two dimensions: time and stations. For consistent testing, we set aside the stations from the VALUE experimental protocol \citep{maraun2015value}. It provides a standardised set of experiments to evaluate downscaling methods and has a specific selection of stations in Europe, of which we select the 53 German stations. These stations cover a wide range of geographic features and are commonly used in downscaling experiments \citep{vaughan2022convolutional}. Because some stations available in VALUE are covered by different copyright permissions than the DWD stations, they are not available to us. In those (9 out of 53 stations) cases, we instead choose the geographically closest DWD station. The furthest discrepancy from the VALUE stations is $\sim \SI{18}{km}$, with most distances below \SI{10}{km}.

\subsection{Training and Validation Stations}
The Train/Val stations are selected in a random order that we then keep fixed so that the stations in the $N_{stations} = 20$ experiments are a subset of the stations used for $N_{stations} > 20$ experiments, which ensures that no information is lost as we increase $N_{stations}$ (Fig. \ref{fig: data splitting} b)). We also investigated choosing stations that are as far apart from each other as possible but found that this significantly hurts what the model can learn, as shorter lengthscale signals are only featured for large $N_{stations}$ in this scenario.

\subsection{Training and Validation Times}
To avoid distribution shift between the Train/Val/Test tasks due to macroscale changes (e.g. climate change, el Niño/la Niña events etc.), we sample times throughout the available range, cycling between 19 days of training data, 2 days of validation data, 2.5 days of testing data, each separated by 2 days that we discard to avoid partial leakage due to correlated tasks, (Fig. \ref{fig: data splitting} c). In this approach, we broadly follow \cite{andrychowicz2023metnet3}, but we increase the number of discarded days from 1 to 2 to further reduce leakage. When we restrict ourselves (artificially) to a limited number of times, we select a random subset of times from the Train/Val pool that we keep fixed across experiments.

\subsection{Generating Tasks}

Once we have selected $N_{stations}, N_{times}$, we try to imitate what we would do for best model performance on downstream applications, with the limited numbers of stations and times available. In such a scenario, we want to maximise the model performance using all $N_{stations}$ available stations.

Given our restricted $N_{stations}$ and $N_{times}$, we follow this procedure (visualised in Fig. \ref{fig: data splitting} a):
\begin{enumerate}
    \item Split $N_{stations}$ into 80\% training and 20\% validation stations.
    \item Split $N_{times}$ into 80\% training and 20\% validation times.
    \item For training, generate random subsets of context and target sets $C_{train}, T_{train}$ only from the set of training stations and times. A single task is drawn as follows:
    \begin{enumerate}
        \item Select a random point in time $t$ from the training times \textit{without replacement} so that each $t$ will be encountered equally often during training.
        \item Draw a fraction $r \sim U(0, 1)$.
        \item Denoting the number of observations at time $t$ as $N_{stations}(t)$, we select a random subset of size $r^2 \times N_{stations}(t)$ as the \textit{context set} $C$. The squaring of $r$ makes sparser tasks more probable, which we find accelerated training for large values of $N_{stations}$.
        \item Use the remaining stations as the target set $T$.
        \item Note that this means a very large number of distinct (but related) tasks can be drawn from a single time $t$, as any combination of context stations is a ``distinct" task. %Throughout this thesis, when we refer to fine-tuning using $N_{times}$ times, we refer to a much larger number of tasks randomly drawn from
    \end{enumerate}
    \item For validation, use \textit{all} available training stations as context $C_{val}$ and all available validation stations as target $T_{val}$ on unseen times from the validation times.
    \item For testing, use \textit{all} available stations, training \textit{and} validation stations as context $C_{test}$, and the test stations (at test times) as targets $T_{test}$. This corresponds to the real-world scenario of using all available stations (train and val) for application. However, this \textit{does} mean the model is tested in a regime that it has not encountered during training (a greater number of context stations).
\end{enumerate}

%\subsection{Spatial or Temporal Validation}
%We also investigated whether or not the model tends to overfit spatially or temporally by qualitatively\footnote{By comparing if the validation loss (held out stations and times) had minima that aligned with spatial validation loss or temporal validation loss.} comparing the validation loss during training to
%\begin{enumerate}
%    \item Spatial validation loss, with stations $\subseteq$ validation stations, but times $\subseteq$ training times,
%    \item Temporal validation loss, with stations $\subseteq$ training stations, but times $\subseteq$ validation times.
%\end{enumerate}
%If overfitting happened to be mainly temporal in nature, it would mean that we would not need a held-out validation set and could train on all $N_{stations}$ stations. However, we found that unless $N_{times} \lesssim N_{stations}$, (which would be very unusual in reality) overfitting was mainly spatial in nature.

\FloatBarrier
\section{Short-Range Artefacts}\label{apx.artefacts}
\begin{figure}
    \centering
    \includegraphics[width=\textwidth]{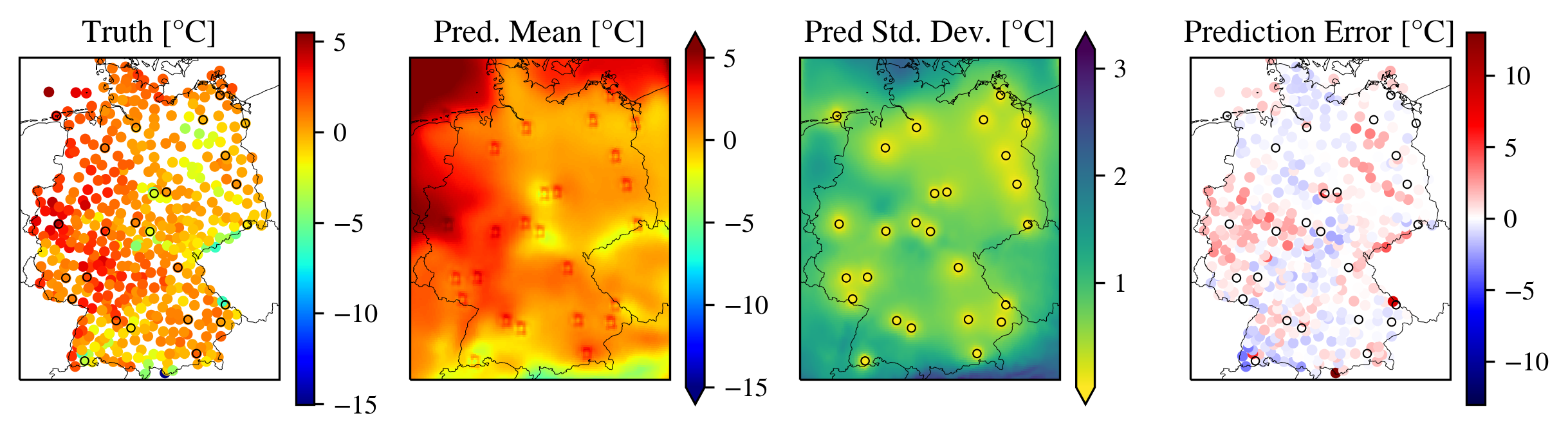}
    \caption{A ConvCNP which has been pre-trained on ERA5 simulator data and not yet fine-tuned, predicting temperature using real context measurements. Immediately around context observations, the model predicts temperatures that clearly do not align with surrounding predictions.}
    \label{fig:artefact prediction}
\end{figure}

A limitation associated with training a ConvCNP using a gridded simulator is the fact that the ConvCNP never receives a training signal shorter than the grid spacing of the simulator, $\ell_{min}$. Attempting to predict at higher resolutions than $\ell_{min}$ can lead to clear visual artefacts (Fig. \ref{fig:artefact prediction}, red smears around context observations). The model is unable to interpolate the context observations smoothly.

These artefacts occur because the model's loss is never punished for predicting them: if the closest separation between two observations is given by $\ell_{min}$, the model's predictions on lengthscales $\ell < \ell_{min}$ have no effect on the loss. The model has not encountered signals of $\ell < \ell_{min}$ in the data and is not endowed with any prior knowledge of temperatures on short scales, it is also unable to extrapolate to shorter lengthscales from the available $\ell > \ell_{min}$ data.

These artefacts are most egregious for Sim Only trained models (where $\ell_{min}$ is comparatively large), but still they remain visible after Sim2Real (Fig. \ref{fig: superresolution}), where $\ell_{min}$ is effectively given by the shortest inter-station separation.

We believe that artefacts are particularly visible because of the U-Net model we use (Fig. \ref{fig: model arch}). The residual connections mean that context observations can pass through the model unmitigated. The single linear layer separating the U-Net from decoding is insufficient for ``smoothing out" the unmitigated artefacts. Expanding the linear layer into a multi-layer perceptron might help mitigate these artefacts, which we believe is worth exploring in future work.

\FloatBarrier
\section{Full Germany Temperature Station Interpolation Sim2Real Experiment Results}\label{apx.full_experiment}
\begin{figure}
    \centering
    \includegraphics[width=\textwidth]{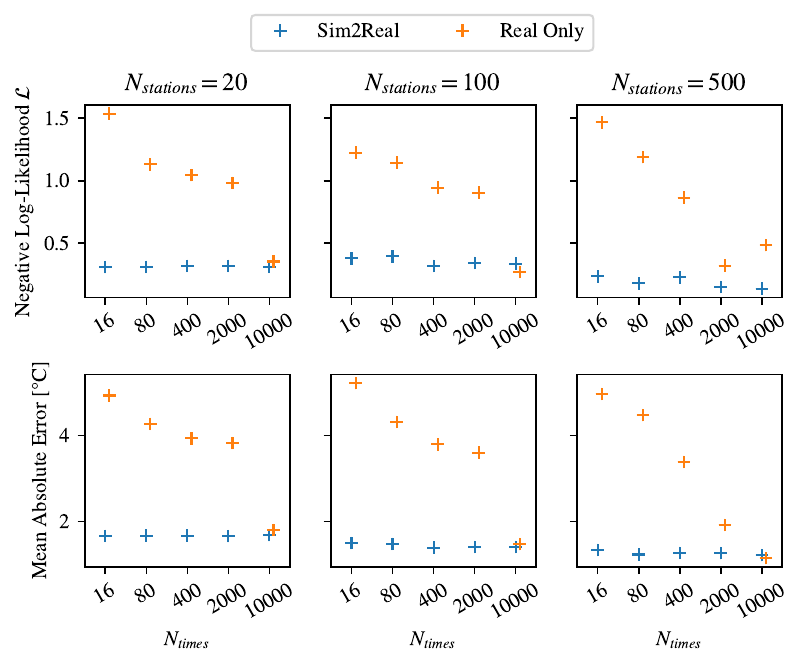}
    \caption{pre-training a model on simulator data significantly aids performance compared to starting from random initialisation. Only with large amounts of real data do the performances become comparable. Note: Fine-tuned model performances (blue) do change with additional training data, but given the scale of the y-axis, this is better shown in Fig. \ref{fig: sim2real vs sim}.}
    \label{fig: sim2real vs real}
\end{figure}

In Fig. \ref{fig: sim2real vs real}, we compare the performance of our Sim2Real transferred model to that trained solely (from random initialisation) on available real data. Clearly, the initialisation at simulator-pretrained parameters is very helpful for training the model in all but the largest real data regime ($N_{times} = 10000$), regardless of $N_{stations}$, showing the utility of Sim2Real in low-data regimes.

\begin{figure}
    \centering
    \includegraphics[width=\textwidth]{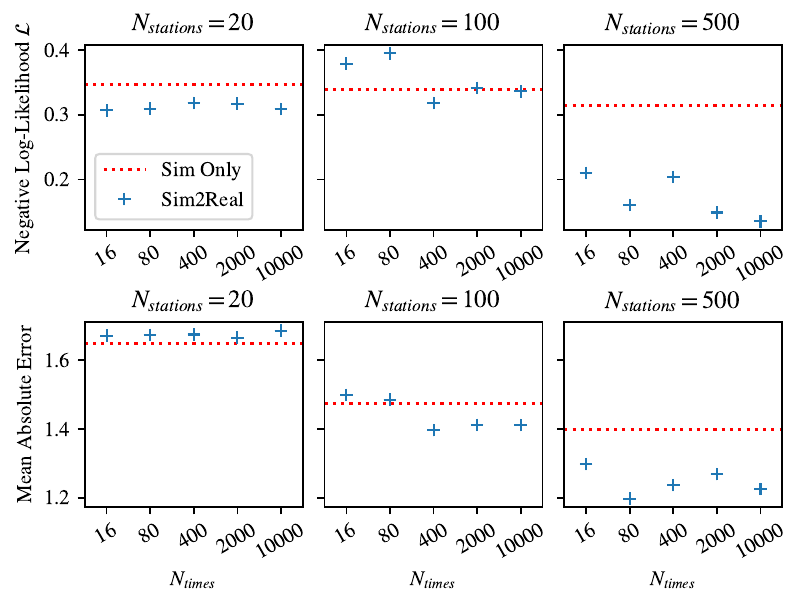}
    \caption{Fine-tuning is only useful when the fine-tuning data is sufficiently different from the simulator. In the $N_{stations} = 20$ regime, the model is unable to improve beyond the simulator, even after many training tasks. In higher data-availability regimes, fine-tuning leads to clear improvements.}
    \label{fig: sim2real vs sim}
\end{figure}

Fig. \ref{fig: sim2real vs sim} shows how the Sim2Real models compare to the Sim Only baseline. We see qualitatively different behaviours in different data availability regimes:
\begin{itemize}
    \item In the sparse-station setup $N_{stations} = 20$, the model is unable to improve through Sim2Real, no matter the quantity of tasks ($N_{times}$).
    \item In the dense-station setup $N_{stations} = 500$, the model does improve significantly, even given very small amounts of real data (e.g. $N_{times} = 16$).
    \item In the middling station density $N_{stations} = 100$, the model does improve slightly given enough tasks $N_{tasks} \gtrsim 400$.
\end{itemize}

These results show that Sim2Real is not always useful, especially when the real data covers a much smaller density of context and target points than the simulator data. However, given even a modestly sized real dataset, Sim2Real can yield significant model improvements.

\end{appendix}

\end{document}